%% file: main.tex
\documentclass[conference]{IEEEtran}
\usepackage{times}
\usepackage{tikz}

\usetikzlibrary{positioning, arrows.meta, calc, backgrounds,quotes, 3d, shapes, arrows, positioning, fit }
\usepackage{booktabs}
\usepackage{multirow}
\usepackage{tabularx}
\usepackage{makecell} 
\usepackage[numbers]{natbib}
\usepackage{multicol}
\usepackage[bookmarks=true]{hyperref}
\usepackage{amsmath}
\usepackage{amssymb}
\usepackage{xcolor}

\usepackage{times}
\usepackage{xcolor}
\usepackage{cleveref}
\usepackage{caption}

\newcommand{\red}[1]{{\color{red}#1}}

\newcommand{\yash}[1]{}
\newcommand{\kar}[1]{}
\newcommand{\sheks}[1]{}

\newcommand{\qq}[1]{Lightning}

\makeatletter
\newif\ifdid@splash \did@splashfalse
\apptocmd{\@maketitle}{%
  \ifdid@splash\else
    \global\did@splashtrue
  \fi
}{}{}
\makeatother

\begin{document}

\title{Adaptive Illumination Control for Robot Perception}

\author{Yash Turkar, Shekoufeh Sadeghi, Karthik Dantu \\ Department of Computer Science and Engineering, University at Buffalo, Buffalo, NY}

\maketitle

\begin{abstract}

Robot perception under low light or high dynamic range is usually improved downstream—via more robust feature extraction, image enhancement, or closed-loop exposure control. However, all of these approaches are limited by the image captured these conditions. An alternate approach is to utilize a programmable onboard light that adds to ambient illumination and improves captured images. However, it is not straightforward to predict its impact on image formation. Illumination interacts nonlinearly with depth, surface reflectance, and scene geometry. It can both reveal structure and induce failure modes such as specular highlights and saturation. These challenges are further exacerbated by robot motion through a scene. \\
We introduce \emph{Lightning}, a closed-loop illumination-control framework for visual SLAM that combines relighting, offline optimization, and imitation learning. This is performed in three stages. First, we train a Co-Located Illumination Decomposition (CLID) relighting model that decomposes a robot observation into an ambient component and a light-contribution field. CLID enables physically consistent synthesis of the same scene under alternative light intensities and thereby creates dense multi-intensity training data without requiring us to repeatedly re-run trajectories. 
Second, using these synthesized candidates, we formulate an offline Optimal Intensity Schedule (OIS) problem that selects illumination levels over a sequence trading off SLAM-relevant image utility against power consumption and temporal smoothness. 
Third, we distill this ideal solution into a real-time controller through behavior cloning, producing an Illumination Control Policy (ILC) that generalizes beyond the initial training distribution and runs online on a mobile robot to command discrete light-intensity levels. Across our evaluation, Lightning substantially improves SLAM trajectory robustness while reducing unnecessary illumination power. We will release our system, code, and datasets upon publication.

\end{abstract}

\IEEEpeerreviewmaketitle

\input{sections/intro}

\input{sections/related}
\input{sections/our-method}

\input{sections/eval}
\input{sections/disc}

\input{sections/conc}


\bibliographystyle{plainnat}
\bibliography{references, misc}

\end{document}

%% file: sections/intro.tex
\section{Introduction}
\label{sec:intro}

\begin{figure}
        \vspace{-0.2cm}
        \centering
        \includegraphics[trim={50pt 20 50 20}, clip, width=\linewidth]{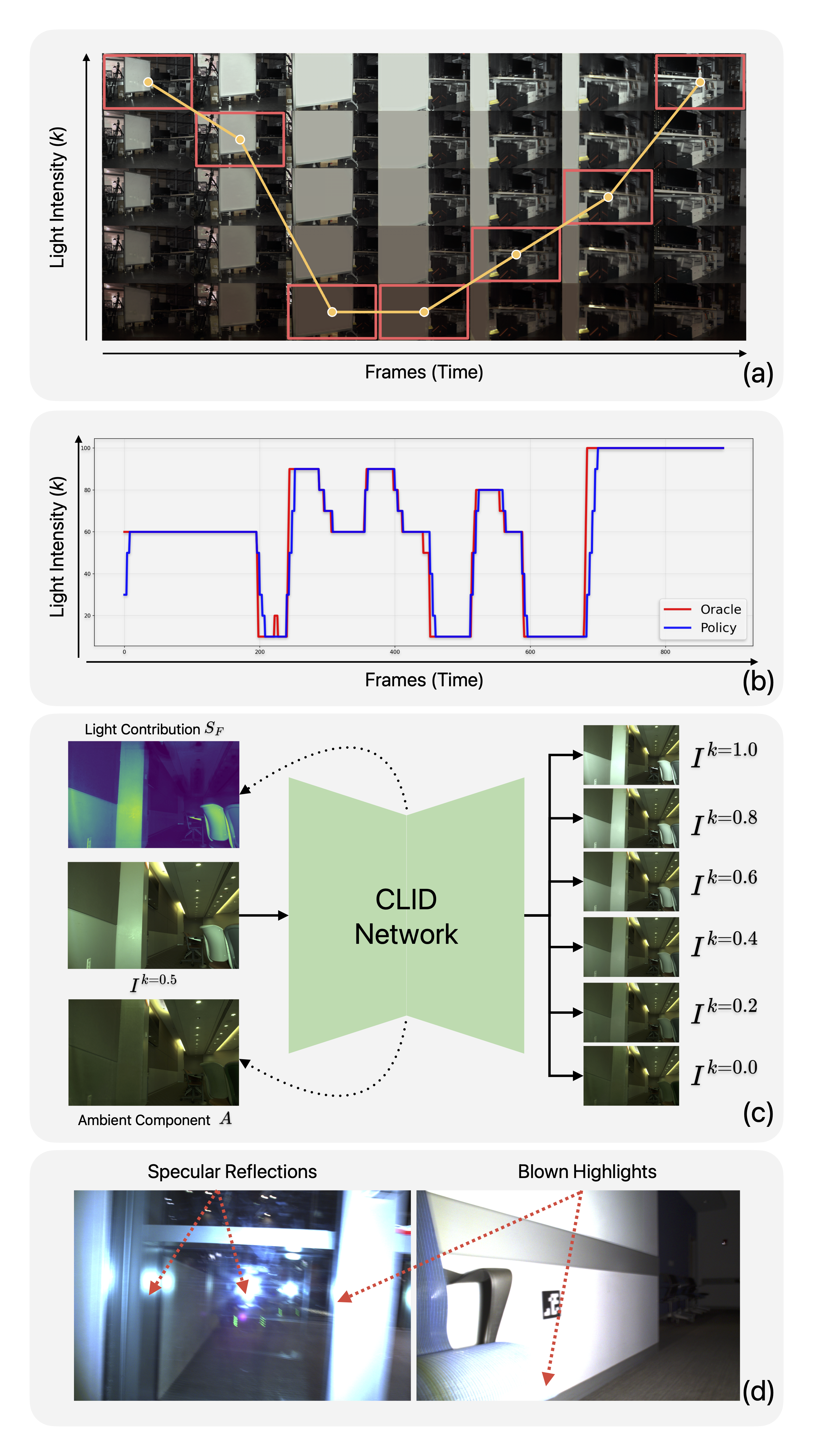}
        \caption{
        \textbf{(a)} Oracle intensity schedule over time (frames vs. discrete levels $K$); orange curve denotes $k^t$ and red boxes mark representative frames. Intensity is reduced near a reflective whiteboard to prevent specularities/saturation, then increased afterward.
        \textbf{(b)} Predicted intensity traces comparing the oracle OIS and learned controller (ILC). 
        \textbf{(c)} CLID relighting: predicted ambient component $A$ and light-contribution map $S_F$ are used to synthesize $I^k$ for candidate intensities $k\in[0,1]$.
        \textbf{(d)} Examples of light sensitivity (specular reflections and blown highlights) motivating adaptive control.}
         \label{fig:intro-splash}
         
\end{figure}

\kar{This needs to be lot more focused upfront - maybe break down what constitutes good image capture. }

Robust perception depends not only on strong perception algorithms, but also on the quality of the underlying images captured. Most innovation in perception has focused on the former, designing algorithms that are robust to noise, nonlinearities, and other harsh real-world effects. While this strategy has been successful in many settings, it is fundamentally limited by the generalization capacity of the algorithms and the variability present in passively collected images.

In contrast, humans and other animals do not rely solely on passive sensing. They actively explore and exploit their environments, using targeted movements and sensor adjustments to acquire more informative observations~\cite{Gibson1983-ln,bajcsy_active_1988,aloimonos_active_1988}. 
In robotics, active perception is often studied as a decision-making problem where robot actions are planned to improve perceptual quality and reduce uncertainty or speed up exploration \cite{bohg_interactive_2017,isler_information_2016}.  
Active perception also encompasses adapting sensor parameters in response to environmental conditions. In this vein, many studies on active exposure control select camera exposure online using scene feedback to improve robustness under difficult lighting~\cite{kim_proactive_2020, gamache_reproducible_2025}.


Environment illuminance strongly conditions robot perception: it sets image brightness and constrains feasible exposure, which in turn determines whether features are detectable and matchable, particularly under rapid or high-dynamic-range lighting variation (e.g., indoor-outdoor transitions, shadows, and low light). Most vision systems rely on auto-exposure (AE) to choose shutter time, gain (ISO), and aperture, but AE is task-agnostic. it regulates global image statistics toward a target mean intensity rather than maximizing task utility, and thus often fails exactly where robustness is most critical. Task-aware active exposure \cite{zhang_active_2017,kim_exposure_2018} closes the loop with the perception objective by selecting exposure to improve signals such as feature count, tracking quality, or detector confidence, but it remains bounded by environment radiance and sensor limits: shutter time cannot increase without sacrificing frame rate or inducing motion blur, and gain cannot increase without amplifying noise. Consequently, when light is insufficient or highly non-uniform, even optimal exposure control cannot recover information that is not present in the measurements.

A natural next step is to move beyond adapting only the sensor parameters and instead act directly on the environment via active illumination \cite{crocetti_comparison_2025}. With an onboard light, the robot (or camera system) can actively shape the lighting field of the scene to make task-relevant structure more observable. This idea already appears in everyday systems such as automotive headlights that automatically adjust beam intensity and direction, flash photography that adds a burst of light to recover detail in dark scenes, and consumer depth cameras that project structured or infrared patterns to enable robust ranging in low light. These systems demonstrate that active illumination can substantially extend the operating envelope of vision systems, but they are typically heuristic, coarse, and only weakly coupled to the actual perception objectives.

Unfortunately, adding artificial light to improve perception is itself non-trivial. The same light that reveals structure in dark regions can also introduce specular highlights, chromatic shifts, and local over-exposure (see \Cref{fig:intro-splash} (d)). These effects can degrade, rather than enhance, the performance of perception algorithms by breaking feature and object tracking, altering appearance, and creating artifacts in the image. In addition, onboard illumination is often power-hungry, and continuous use can significantly reduce a robot’s operational time. Therefore, it is important to use onboard light selectively and intelligently to explicitly improve the perception objective. 

\begin{figure*}
    \centering
    \includegraphics[trim={1cm 1cm 1cm 1cm}, clip,width=\linewidth]{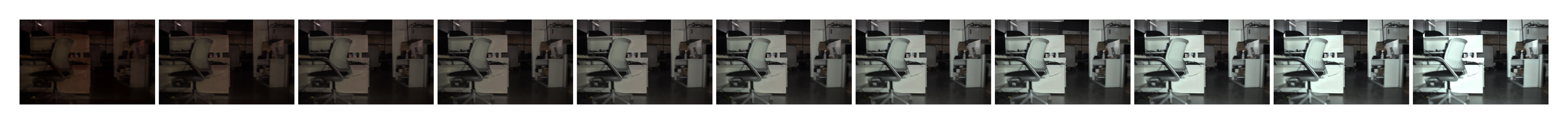}
    \caption{\textbf{CLID relighting on robot sequences.} A random frame from sequence \texttt{lab2113} shown after relighting. Raw images are captured by setting light intensity to 50\% $(k=0.5)$, these are then relit using the CLID network to generate images at light-intensities 0-100\% at 10\% increments (left to right) 
    }
        \label{fig:clid-sequence}
        \vspace{-0.2cm}
\end{figure*}

Building on this view, we ask a concrete question: \textit{Given a robot equipped with an onboard, co-located light source, how should the robot illuminate its surroundings to improve perception?} "Good perception" depends on the downstream task. We focus on visual SLAM, and formulate a sequence-level decision problem: along a fixed robot trajectory, choose a time-varying sequence of light intensities that maximizes this objective. Such a sequence is called the Optimal Intensity Schedule (OIS). 

Computing the OIS is difficult because it requires images of the same scene under every candidate intensity at every time step $t$. This is tedious and impractical to capture in real-world settings. We overcome this challenge by learning a relighting model that takes an  observed frame at given light intensity and a target light intensity, and synthesizes the view under that illumination. This model lets us convert a recorded run into a dense, virtual sequence spanning all candidate intensities (see \Cref{fig:clid-sequence}). Over this expanded sequence, we cast the OIS selection as an energy minimization problem \cite{koller_probabilistic_2009,wainwright_graphical_2008,szeliski2022cv} and solve it offline using dynamic programming, producing an \emph{oracle} that yields high-quality schedules for arbitrary recorded runs.

Since the oracle requires access to future frames, it cannot be executed online. Instead, we use it to supervise a learned controller, an imitation learning policy (ILC) that conditions on the current image and the previous light setting to predict the next intensity from a discrete set of $\mathcal{K}$ levels. This yields a task-aware active illumination strategy that approximates oracle behavior in a real-time robot deployment.

The contributions of our work are as follows:
\begin{itemize}
    \item A co-located illumination decomposition (CLID) and relighting model that enables synthesizing scene appearance under arbitrary onboard light intensities from sparse observations.
    \item A sequence-level formulation of optimal intensity scheduling (OIS) as energy minimization on the relit sequences 
    \item A real-time active illumination controller (ILC) trained via imitation learning from the oracle, which predicts the next discrete intensity level from visual input and prior actuation.
\end{itemize}

In this paper, we focus exclusively on \emph{active illumination control} with \emph{fixed camera exposure}. We treat onboard light intensity as the primary control input and hold exposure constant to isolate the effect of illumination on perception. Jointly optimizing illumination and exposure is future work, and we note that there are complementary methods that could be integrated into \qq{} for this purpose. 



 This paper is organized as follows: \Cref{sec:related} details related work. \Cref{subsec:relighting} explains our relighting network architecture while \Cref{subsec:oracle} describes how we compute the optimal intensity schedule, finally, \Cref{subsec:policy} covers our imitation learning policy. \Cref{sec:eval} shows offline and real-world experimental results and in \Cref{sec:discussion} we discuss our experimental insights and future directions of this work.

%% file: sections/related.tex
\section{Related Work}
\label{sec:related}


Relevant prior work spans three areas: (i) SLAM/VO techniques that improve robustness under low light and photometric variation, (ii) active exposure and illumination control and (iii) Image relighting methods.


\subsection{Robust SLAM in low-light environments}


Robustness in SLAM has been improved via both front-end design and photometric modeling. Feature-based systems such as ORB-SLAM3 \cite{campos_orb-slam3_2021} leverage robust tracking, place recognition, and multi-map reuse to recover from tracking loss across diverse environments. Direct methods (e.g., LSD-SLAM \cite{engel_lsd-slam:_2014} and DSO \cite{engel_direct_2016}) minimize photometric error and can be accurate, but are sensitive to brightness-consistency violations from auto-exposure, vignetting, and nonlinear camera response. DSO mitigates this with an explicit photometric model (exposure, vignetting, response), motivating online photometric calibration \cite{bergmann_online_2017} that estimates per-frame exposure and response (and vignetting) to enable direct VO/SLAM on auto-exposure videos without specialized calibration. However, such methods remain limited in photon-starved settings where low signal yields insufficient stable features or gradients.

\subsection{Active exposure and illumination control}
A variety of approaches to automatic exposure control have been introduced in prior work~\cite{zhang_image_2024,han_camera_2023,zhang_efficient_2025,kim_exposure_2018,mehta_gradient-based_2020,lee_learning_2024,gomez-ojeda_learning-based_2018,liu_learning-based_2020,tomasi_learned_2021}. 
While these methods tune imaging parameters such as exposure-time (shutter speed) and gain (ISO) - e.g., regulating exposure-time~\cite{liu_learning-based_2020} or jointly adjusting exposure time and sensor gain~\cite{tomasi_learned_2021}; their effectiveness can be limited in low-light settings or under rapid illumination changes. 
In such conditions, longer exposures or higher gain are often required to obtain usable images, which can reduce the effective frame rate and amplify noise.

Motivated by these limitations, some approaches instead augment scene illumination using an onboard light source. 
This is a promising direction, but \cite{crocetti_comparison_2025} show that naive illumination control can degrade performance due to specular reflections and can also waste valuable onboard power. 
NightHawk~\cite{turkar_active_2025} performs active exposure and illumination control via event-triggered Bayesian optimization. However, its computations require the robot to stop and calculate the illumination setting which prevents fully real-time operation. 
Finally, \cite{crocetti_active_2025} adjusts the direction of the light to illuminate regions with strong features, but does not regulate light intensity.

\subsection{Image relighting}

Image relighting has a long history in computer graphics and vision, where the goal is to predict how scene appearance changes under novel illumination. Early approaches showed that relighting can be achieved by explicitly capturing illumination and rendering with physically motivated models \cite{debevec_rendering_nodate}, but they typically assume controlled acquisition, calibration, and/or specialized hardware requirements that are difficult to satisfy in in-the-wild settings. More recently, learning-based relighting methods \cite{einabadi_deep_2021, bhattad_stylitgan_2024, xing_luminet_2024} have framed relighting as a conditional prediction problem: given an input image, predict a relit output conditioned on target lighting parameters. Representative examples include \cite{zhang_latent_2025, xing_luminet_2024}, which disentangle lighting and intrinsic content in a latent space and relight scenes by transferring illumination from a reference. \cite{maralan_computational_2023} targets flash/ambient relighting by estimating flash and ambient shading, but requires depth and image intrinsics as additional inputs obtained from other procedures. \cite{bell_intrinsic_2014} proposes an intrinsic decomposition and introduced one of the first datasets designed to support learning-based relighting and decomposition.



%% file: sections/our-method.tex
\section{Our Approach}
\label{sec:our-method}

\begin{figure*}
    \centering
    \includegraphics[width=\linewidth]{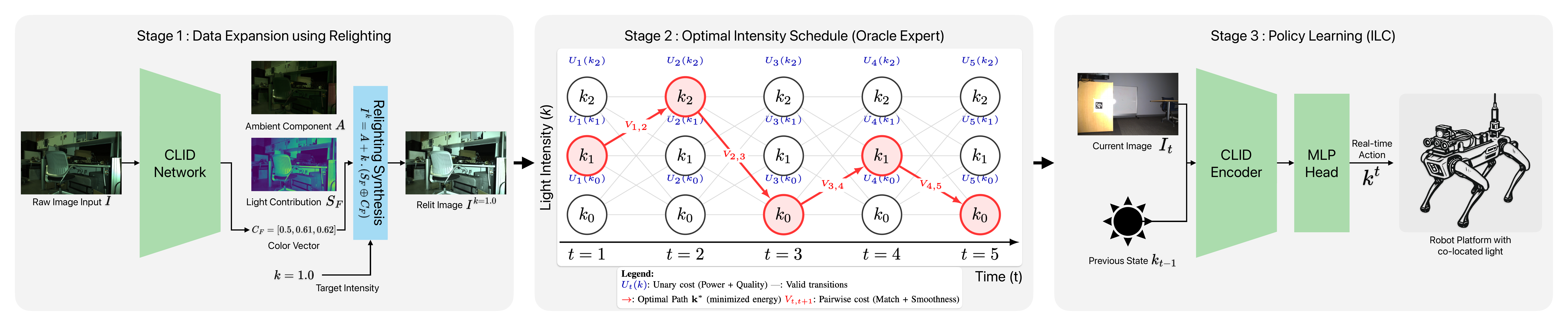}
    \caption{\textbf{\qq{} Framework:} Our pipeline consists of three  stages: 
    \textbf{(1) Data Expansion (Relighting):} A Co-Located Illumination Decomposition (CLID) network takes a real observation at some light-intensity (50\% for our experiments) and decomposes it into an ambient component $A$, a scalar light contribution map $S_F$, and a color vector $C_F$. This allows for the synthesis of a dense virtual dataset spanning the full range of candidate light intensities $\mathcal{K}$.
    \textbf{(2) Expert Generation (Oracle):} As illustrated in the trellis graph (Stage 2), we formulate active illumination as a global optimization problem. The optimization identifies the Optimal Intensity Schedule (OIS) by traversing a trellis structure where nodes represent discrete light levels and edges represent transitions. The path is determined by minimizing a sequence-level energy function comprising unary costs (balancing power consumption and image utility) and pairwise costs (balancing feature matching quality and temporal smoothness). 
    \textbf{(3) Imitation Policy (ILC):} The offline Oracle is distilled into a real-time policy via behavior cloning. The resulting controller uses a visual encoder from CLID to process the current image $I_t$ and previous light state $k_{t-1}$ to predict the next optimal illumination intensity $k_t$ for online deployment.}
    \label{fig:full-arch}
\end{figure*}

 Our approach to active illumination control follows a three-stage pipeline shown in \Cref{fig:full-arch}. First, we train an image relighting network (\Cref{subsec:relighting}) that takes an image captured under a non-zero light intensity and synthesizes a dense set of relit images spanning the full range of candidate intensities (0\%--100\% at 10\% increments). \Cref{fig:clid-sequence,fig:intro-splash} show examples of the resulting synthesized dataset. Second, we use this dataset to compute an optimal intensity schedule offline via dynamic programming, which we refer to as the oracle method (\Cref{subsec:oracle}). The schedule shown in \Cref{fig:intro-splash}(a) selects the best light intensity at each time step while also accounting for power and smoothness of operation. 
Finally, we use the oracle-derived schedules across many sequences to train an imitation-learning policy (\Cref{subsec:policy}) that generalizes across sequences and runs online, enabling real-world deployment when future observations and relit data are not available at run time.

\input{sections/relighting}

\input{sections/oracle}
\input{sections/lightning}

%% file: sections/relighting.tex
\subsection{Relighting Network}
\label{subsec:relighting}

Our relighting task builds on the principle of intrinsic image decomposition \cite{Barrow1978RECOVERINGIS,barron_shape_2020, narihira_direct_2015}, which models the image in terms of reflectance and shading:

\begin{equation}
    I = R\ .\ S
    \label{eq:iid}
\end{equation}

where $I$, $R$ and $S$ represent the input image, reflectance (albedo) and shading respectively. Considering a co-located light or flash, \cite{maralan_computational_2023} defines image formation as:

\begin{equation}
    I = R\ .(c_AS_A + S_F)
    \label{eq:marlan-flash}
\end{equation}

where $S$ is split into ambient $S_{A}$ and flash $S_{F}$ shading while $c_A$ is a 3-dimensional vector for color temperature of the ambient light 

\Cref{eq:marlan-flash} is possible because, an image captured with a co-located light, like a camera flash can be modeled using the illumination superposition principle, which is the summation of ambient and co-located light illuminations (flash) in linear RGB space:

\begin{equation}
    I = A + S_{F}
    \label{eq:image-superposition}
\end{equation}

where $A$ and $S_{F}$ represent the ambient illumination ($R\ .\  S_{A}$) and the flash shading respectively. 









We build on \Cref{eq:iid,eq:image-superposition,eq:marlan-flash} and define a relaxed form of intrinsic image decomposition which we term \emph{co-located illumination decomposition} (CLID). Rather than recovering physically accurate reflectance and shading under arbitrary lighting, we consider a co-located camera-light setup (like a flash \cite{petschnigg_digital_nodate,eisemann_flash_nodate, nayar_fast_nodate}) and decompose an image captured with non-zero flash intensity into (i) an ambient component and (ii) a co-located light / flash contribution. These components then enable synthesis under arbitrary light intensities via

\begin{equation}
\label{eq:relight}
    I^{k} = \bigl(A + k \cdot (S_{F} \oplus C_F)\bigr), \qquad k \in \mathcal{K} \subseteq [0,1].
\end{equation}

where $I^{k}$ is the synthesized image at light intensity $k$, $A \in \mathbb{R}^{H \times W \times 3}$ is the ambient component, $S_F \in \mathbb{R}^{H \times W \times 1}$ is a co-located light contribution map, and $C_F \in \mathbb{R}^{1 \times 3}$ is a global color vector that captures the chromaticity of the co-located light. In contrast to \cite{maralan_computational_2023}, we do not assume a white-balanced flash image or an achromatic light, rather we use $C_F$ to encode the light's color. The scalar $k$ controls the co-located light's intensity.

It is important to note that $A$ is not a true albedo or reflectance map in the classical intrinsic-image sense (see \Cref{fig:intro-splash,fig:full-arch}). Instead, it is a lumped “ambient image” that contains albedo and ambient shading, and is intended to match the scene if captured without the co-located light under the same ambient illumination and exposure. \Cref{eq:relight} is defined in linear sensor space, i.e.,  all operations are applied directly to \texttt{BayerRG16} RAW mosaiced images before the camera response function (CRF) and white balance, and under a fixed exposure setting at capture time.

\subsubsection{Network Architecture}
\label{subsec:architecture}
To estimate the components $A$, $S_F$, and $C_F$, we use a standard U-Net-style encoder--decoder architecture \cite{1505.04597}. The network takes a raw RGB image $\mathbf{I}\in\mathbb{R}^{H\times W\times 3}$ and a scalar light intensity $k\in\mathbb{R}$ (4 channel input) and processes it through a multi-scale convolutional backbone with skip connections, which encodes global context at coarser resolutions and decodes back to full resolution while preserving spatial detail.

At the final decoder resolution, we attach a single $1 \times 1$ convolutional layer that produces a 4-channel output map
$Y \in \mathbb{R}^{H \times W \times 4}$ where the first three channels are the ambient RGB map and the remaining channel as the scalar light map.
Thus $A$ and $S$ are obtained by a simple channel-wise split of the shared prediction tensor.
The color vector $C_F \in \mathbb{R}^3$ is implemented as a small set of learnable parameters that is optimized jointly with all network weights. 

\subsubsection{Training and Supervision}
\label{subsec:training}

We train CLID in two stages. We first pre-train on the MIT Multi-Illumination \cite{murmann_dataset_2019} dataset, which contains images of scenes with lights from various directions, we treat direction $0$ as ambient and direction $2$ as the co-located source, and synthesizing intermediate intensities to supervise relighting. We compute total loss ($\mathcal{L}_{\text{total}}$) as
\[
\mathcal{L}_{\text{total}}
=\lambda_{\text{rec}}\mathcal{L}_{\text{rec}}
+\lambda_{\text{amb}}\mathcal{L}_{\text{amb}}
+\lambda_{\text{aux}}\mathcal{L}_{\text{aux}}
\]
where $\mathcal{L}_{\text{rec}}$ and $\mathcal{L}_{\text{amb}}$ are reconstruction losses for the synthesized and ambient image, and $\mathcal{L}_{\text{aux}}$ collects auxiliary consistency regularizers. During pretraining we set $\lambda_{\text{zero}}=0$, where $\lambda_{\text{zero}}$ weights the \emph{zero-intensity} auxiliary loss component $\mathcal{L}_{\text{zero}}$ (flash off: $k=0$, so the $S_F$ should vanish and the image reduces to $A$). We then fine-tune on our captured intensity dataset with $\lambda_{\text{zero}}>0$ to anchor ambient effects in $A$ and prevent leakage into the light component. Implementation and dataset details are deferred to \Cref{sec:eval}.

%% file: sections/oracle.tex
\subsection{The Oracle}
\label{subsec:oracle}


Using the trained relighting network, we synthesize images for all candidate illumination settings, i.e $k=[0.0, 0.1,\dots,1.0]$ (0\% to 100\% in 10\% increments), for each observed image during robot traversal. This expanded dataset enables an offline formulation of active illumination as a \textbf{discrete energy minimization} problem over a sequence of discrete variables: among all feasible intensity assignments, we seek the one that optimizes task performance while satisfying power and smoothness constraints. We refer to the optimal solution of this offline problem as the \textit{Oracle}.





\subsubsection{Problem Setup}

Consider a recorded sequence of length $T$ with a known robot trajectory. Using the relighting network, we generate a dense virtual dataset in which, for each time step $t \in \{1,\dots,T\}$ and each candidate light intensity $k \in \mathcal{K}$, a synthesized image $I_t^{k}$ is available. Here, $\mathcal{K} \subset [0,1]$ denotes a finite set of physically realizable onboard light intensities (e.g., $\mathcal{K} \subset [0.0,0.1,\dots,1]$, uniformly spaced between 0\% and 100\% power at 10\% increments).

An \textit{intensity assignment} is defined as a sequence of intensity values
\begin{equation}
    \mathbf{k}_{1:T} = (k_1, \dots, k_T), \quad k_t \in \mathcal{K},
\end{equation}
which uniquely specifies the onboard light intensity at each time step. Our goal is to compute the \emph{Optimal Intensity Schedule} (OIS), denoted $\mathbf{k}^*_{1:T}$, that minimizes a sequence-level energy capturing task performance over the entire trajectory while accounting for power consumption and penalizing rapid intensity changes. In the remainder of this paper, we instantiate task performance using SLAM/VO metrics, and infer the corresponding intensity schedule.



\subsection{Sequence-level Cost}

We define a sequence-level energy function $E(\mathbf{k}_{1:T})$, where lower energy corresponds to a more desirable assignment:
\begin{equation}
    E(\mathbf{k}_{1:T}) = \sum_{t=1}^{T} U_t(k_t) + \sum_{t=1}^{T-1} V_{t,t+1}(k_t, k_{t+1}).
    \label{eq:global_energy}
\end{equation}


The oracle corresponds to the globally optimal intensity schedule
\(\mathbf{k}^*_{1:T} = \arg\min_{\mathbf{k}_{1:T}} E(\mathbf{k}_{1:T})\).
The energy decomposes into unary potentials \(U_t(\cdot)\), which evaluate the per-frame desirability of selecting a particular intensity, and pairwise potentials \(V_{t,t+1}(\cdot,\cdot)\), which capture inter-frame effects such as feature matching quality and temporal smoothness.

\paragraph{Unary Terms (Power and Image Quality)}
The unary potential is defined as a weighted sum of image quality penalties and power consumption:
\begin{equation}
    U_t(k) = \lambda_{D} D_t(k) + \lambda_{P} P(k).
\end{equation}
Here, $P(k)$ models the power cost of using intensity level $k$ (a monotonic function $P(k) \propto k$). The term $D_t(k)$ is a heuristic penalty on the quality of the relit image $I_t^{k}$, capturing factors such as uneven brightness and lack of texture. 

The non-negative weights $\lambda_{D}$ and $\lambda_{P}$ govern the influence of these local constraints. For instance, increasing $\lambda_{P}$ biases the oracle toward lower intensity schedules to conserve energy. Setting $\lambda_{D} = 0$ removes the reliance on heuristic image metrics, yielding a task and power-driven oracle that optimizes solely for the end-to-end SLAM metric (defined in the pairwise term) and power efficiency.

\paragraph{Pairwise Terms (Feature Matching and Smoothness)}
For consecutive time steps $(t, t+1)$ and candidate intensities $(k, l)$ ($l$ is light intensity at t+1), we define a pairwise potential that balances task performance with temporal consistency:
\begin{equation}
    V_{t,t+1}(k,l) = \lambda_{M} \cdot (1 - M_{t,t+1}(k,l)) + \lambda_{S} \cdot|k - l|.
\end{equation}
The matching score $M_{t,t+1}(k,l)$ measures the utility of the image pair $(I_t^{k}, I_{t+1}^{l})$ for the SLAM/VO system. In a feature-based system, $M$ represents the ratio of feature-match inliers. The term $\lambda_{M} (1-M_{t,t+1}(k,l)$) (where $\lambda_{M} > 0$) assigns lower energy for intensity pairs that yield high-quality measurements.

The term $\lambda_{S} \cdot |k-l|$ (where $\lambda_{S} > 0$) penalizes large frame-to-frame intensity jumps. This encourages temporally smooth lighting, implicitly preventing flickering.

\subsubsection{Dynamic Programming Solution}

The energy function in \Cref{eq:global_energy} possesses a chain structure, equivalent to a linear-chain Markov Random Field (MRF). Global minimization of such chain-structured energies can be solved exactly via min-sum dynamic programming on linear chains.  \cite{koller_probabilistic_2009} \cite{wainwright_graphical_2008}\cite{szeliski2022cv}. We define the optimal prefix energy $F_t(k)$ as the minimum energy of all partial assignments up to time $t$ ending in intensity $k$:
\begin{equation}
    F_t(k) = \min_{k_{1:t-1}} E(k_{1:t-1}, k_t = k).
\end{equation}
The recurrence relation proceeds as follows:
\begin{enumerate}
    \item \textbf{Initialization} ($t=1$): 
    \begin{equation}
        F_1(k) = U_1(k).
    \end{equation}
    \item \textbf{Forward Pass} ($t=2, \dots, T$): For each intensity $k$,
    \begin{equation}
        F_t(k) = U_t(k) + \min_{l} \left[ F_{t-1}(l) + V_{t-1,t}(l,k) \right].
    \end{equation}
    We store the optimizing index $l$ as a backpointer for retrieval.
    \item \textbf{Backtracking}: The optimal final intensity is $k_T^* = \arg\min_k F_T(k)$, and the full assignment $\mathbf{k}^*_{1:T}$ is recovered by tracing the backpointers from $T$ to $1$.
\end{enumerate}
Given $|\mathcal{K}|$ discrete light levels (typically small, e.g., $10$)
, the complexity is $O(T|\mathcal{K}|^2)$, which is negligible compared to the cost of computing the matching scores $M_{t,t+1}$.


%% file: sections/lightning.tex
\subsection{Illumination Control Policy (ILC)}
\label{subsec:policy}




The oracle is computed offline by optimizing over the expanded relit dataset, which provides images for every candidate intensity at every time step. Realizing this procedure online would require generating candidate relit views and evaluating image-utility and matching costs on the fly, while also losing access to future frames used by the oracle. This makes direct online deployment impractically expensive. We therefore distill the oracle into a lightweight imitation policy via behavior cloning, enabling real-time active illumination control.

As shown in \Cref{fig:full-arch}, the policy takes as input the current image $I^t$ and the previously commanded light level $k_{t-1}$, and predicts the next discrete intensity $k_t \in [0,1]$. We encode $I_t$ using a CLID-pretrained visual encoder to extract high-level lighting-relevant features, and concatenate this embedding with $k_{t-1}$. A small MLP head maps the resulting feature vector to logits over the $\mathcal{K}$ intensity levels. We train on oracle OIS sequences using a cross-entropy loss between the predicted logits and the oracle action. \\ 


\noindent\textbf{Temporal-rate mismatch:}
The oracle OIS is computed at the dataset rate (10\,Hz), but commanding the onboard light at this frequency is infeasible due to hardware and control-loop limitations. Naïvely training on adjacent 10\,Hz transitions $(I^t, k^*_{t-1}) \mapsto k^*_t$ induces a distribution shift at deployment, where actions are held for multiple frames and the policy observes larger temporal gaps between decisions. To align training with the deployment control rate, we construct temporally subsampled supervision by forming training tuples at multiple strides $\Delta$:

\begin{equation}
(I^t,\; k^*_{t-\Delta}) \mapsto k^*_t,\qquad \Delta \in \mathcal{D}.
\end{equation}

In practice, for 1\,Hz control we include $\Delta=10$ (and nearby offsets for robustness), yielding pairs such as $(1\!\rightarrow\!10)$, $(2\!\rightarrow\!11)$, etc. Training on this mixture improves performance under slower actuation and scheduling jitter by matching the time-pairs encountered at inference.

%% file: sections/eval.tex
\section{Experimental Evaluation}
\label{sec:eval}

\input{tables/relight-static}

In this section we describe our experimental setup, evaluation methods and results of each process.

\subsection{Datasets, data-collection, robot setup}
\label{subsec:setup}

\begin{figure}
    \centering
    \includegraphics[width=\linewidth]{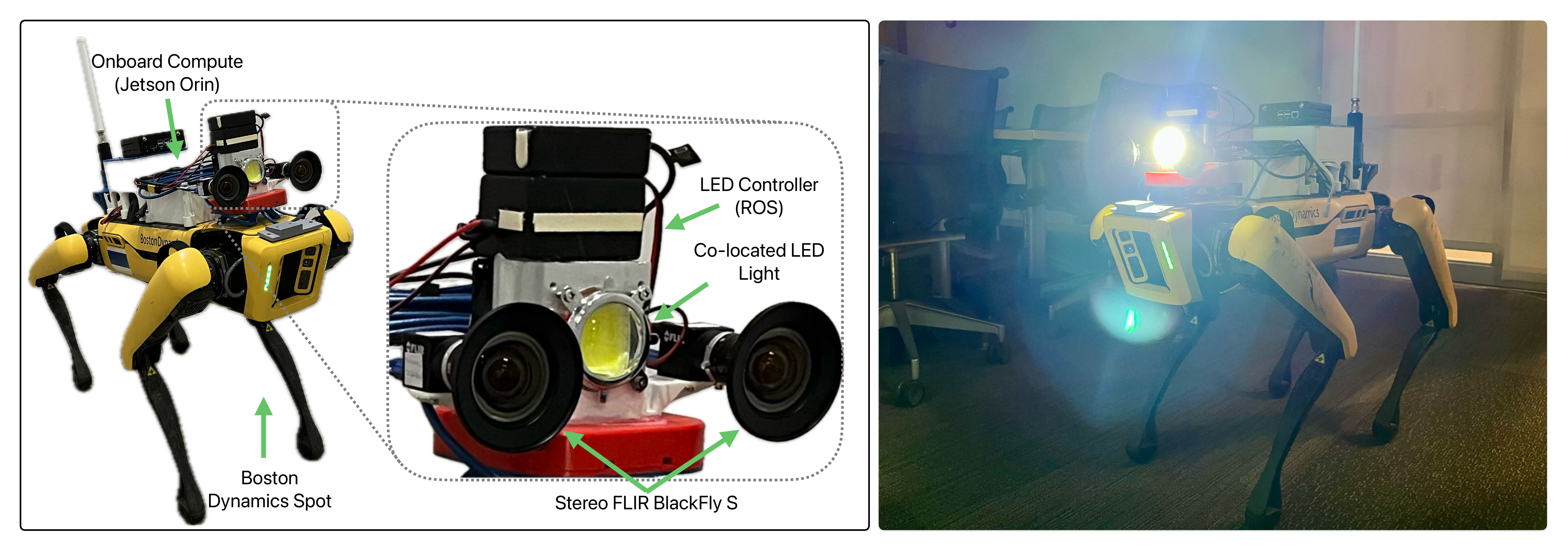}
    \caption{Our robot platform for data-collection and evaluation. \textbf{Left}, an illustration of our hardware including the stereo camera setup and custom-built co-located light and \textbf{right} shows our robot operating in a low-light environment}
    \label{fig:hardware}
\end{figure}

For pre-training the relighting network, we use the MIT Multi-Illumination Dataset \cite{murmann_dataset_2019} which contains images of 1016 indoor scenes captured under 25 distinct illumination directions sampled over the upper hemisphere. The dataset provides scene-linear images in EXR format, enabling physics-based relighting operations. We select illumination directions 0 and 2, which we treat as ambient and co-located lighting conditions, respectively. Intermediate intensity images are synthetically generated between these two conditions to simulate continuous variations in light intensity from a fixed direction.

We additionally collect still scenes and image sequences for SLAM/VO using a custom camera and co-located lighting setup mounted on a mobile robot. 
Images are captured using a FLIR Blackfly S camera with fixed exposure settings so that the images are neutrally exposed. The co-located light source is a custom-built COB LED equipped with 120° FOV (see \Cref{fig:hardware}).
For sequence capture, two cameras are hardware-synchronized at 10Hz. All images are recorded in the $\textrm{BayerRG16}$ format, i.e., RAW sensor measurements without demosaicing, white balance, or gamma correction, which preserves the scene-linear image formation model \Cref{subsec:relighting}. The sensing payload is mounted on a Boston Dynamics Spot robot, enabling reliable and repeatable teleoperated and autonomous data collection. The robot’s autonomy package provides ground-truth pose estimates via a multi-camera and fiducial-based localization system, which we use as a reference for SLAM/VO evaluation.

We collect 160 static scenes for training the relighting network, capturing each scene at 25\% increments of co-located light intensity. We further record 10 sequences using the custom camera-light setup. Each sequence is acquired via autonomous navigation at 50\% light-intensity. These are then used as input to the relighting network to synthesize a multi-intensity dataset, yielding a dense set of images spanning 0–100\% light intensity in 10\% increments for each time step. We also capture with 0\% and 100\% settings for 7 out of the 10 sequences to evaluate relighting performance.

SLAM performance is evaluated using weighted RMSE (WRMSE) (similar to \cite{crocetti_comparison_2025}), this penalizes sequences where the predicted trajectory is of different length compared to reference.

\input{tables/oracle-eval}

\subsection{Relighting Performance}
\label{subsec:relighting_eval}


We evaluate relighting fidelity in two settings: (i) our custom-captured static scenes, and (ii) the real robot sequences. For (i), paired reference images at known light intensities are available; we therefore quantify reconstruction quality using PSNR and SSIM \cite{1284395}, which measure pixel-wise accuracy and structural similarity, respectively. Representative results are provided in \Cref{tab:relight_fidelity_paired_meanstd_singlecol} where the synthesized images closely match the target illumination and preserve fine-scale appearance, indicating strong relighting performance. We additionally observe that the predicted light-contribution heatmaps ($S_F$) are intuitive and consistent with the underlying scene geometry, highlighting regions that are most affected by changes in illumination.

For sequences, exact frame-to-frame pose correspondence or temporal alignment across different lighting intensity runs is not guaranteed; therefore, per-frame supervised evaluation metrics are not directly applicable. Instead, we exploit the fact that each trajectory is also captured at reference lighting intensities of $0\%$ and $100\%$. We compare the synthesized intensity sequences against these references by performing feature detection and reporting the percentage change in (i) the mean number of detected keypoints and (ii) the mean image intensity. High agreement in these statistics indicates that the relighting process preserves salient visual features.


Across sequences, the synthesized images exhibit a mean percentage change in the number of detected keypoints of $-20.1\%$ relative to the $0\%$ reference, while the performance is substantially better for the $100\%$ reference, with a change of only $0.1\%$. The mean image intensity change follows a similar trend: the $0\%$ case shows a change of $-15.5\%$ indicating that the relit images are slightly dimmer, whereas the $100\%$ case shows a change of $8.7\%$, indicating slightly brighter relit images. Overall, these values are averaged over multiple sequences and demonstrate that the network achieves adequate reconstruction performance. Qualitative results are presented in \Cref{fig:clid-sequence}, where a random frame is shown at different lighting intensities (x-axis, increasing from $0\%$ to $100\%$).

\subsection{Optimal Intensity Schedule (OIS) using Oracle}
\input{tables/rollout}

The oracle computes the OIS, i.e., a discrete light-intensity value at each time step that maximizes visual SLAM performance. In our experiments, we instantiate the matching score $M$ using MASt3R~\cite{leroy_grounding_2024}, which produces feature and geometry-based correspondences via a fast nearest-neighbor matcher. We normalize the raw number of unique reciprocal MASt3R matches by the theoretical maximum number of sampled grid locations (i.e., the maximum possible reciprocal matches for the chosen subsample), yielding a matching score in ([0,1]) for the optimizer. For image utility, we use $M_{feat}$ from \cite{turkar_active_2025} which is already normalized between ([0,1]).


The scalar weights in our oracle cost are tuned with Optuna~\cite{1907.10902} (Bayesian optimization). Using these tuned weights, we run the oracle to obtain a per-sequence OIS and compare it against constant 0\% and 100\% light-intensity baselines. To ensure fairness, all SLAM evaluations use the relit dataset. We report  trajectory-ratio $C$, WRMSE, mean light-intensity and power usage $P$ in \Cref{tab:ois_mast3rslam}.

It is evident that the OIS outperforms all baselines every sequence in terms of WRMSE. In the \texttt{113through} sequence, which begins with a strong dark-to-bright transition accompanied by reflections, both the 0\% and 100\% baselines fail, as indicated by their trajectory ratios $C$, whereas the oracle is able to track for a substantially longer duration (approximately 50\%). Similarly, in \texttt{entry2corr}, which contains a transition event in which the robot observes a reflective metal door, both baselines perform worse than the oracle. In \texttt{labin}, both baselines fail at $C=0.5$ while the oracle is able to complete the trajectory.
These trends are consistent across most sequences, where the oracle can be observed actively adjusting light intensity to compensate for reflections, over-exposure, and ambient lighting changes (see supplementary material for qualitative examples).

\begin{figure*}
    \centering
    \includegraphics[width=\linewidth,trim={20pt 40 40 20},clip]{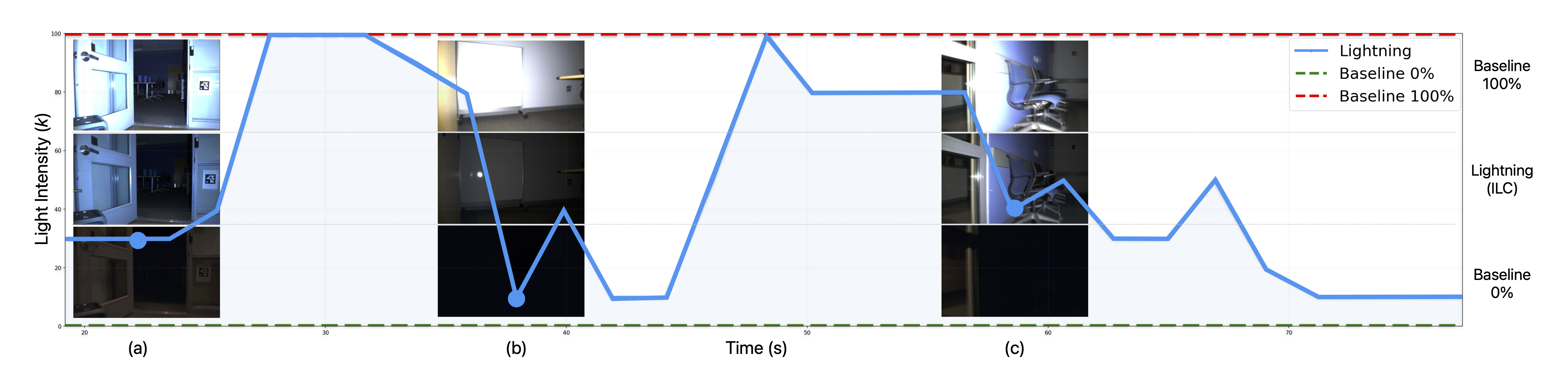}
    \caption{\emph{ILC's intensity schedule versus fixed baselines}. The imitation policy, deployed on a robot, outputs a per-frame light intensity (blue), and is compared against 0\% (green-dashed) and 100\% (red-dashed) fixed-intensity baselines. Insets show frames at the corresponding timestamps: (a) ILC increases illumination when entering a low-light region to preserve image utility; (b) ILC reduces illumination near a reflective whiteboard to mitigate specular saturation; (c) ILC chooses an intermediate illumination level to balance competing effects of low light and specular reflection.\yash{fix explanation}}
    \label{fig:placeholder}
\end{figure*}

\subsection{Active Illumination with ILC}

We train an imitation-learning policy, ILC,  using the oracle experts. ILC is trained on 4 x NVIDIA H200s for 10 hours and the final model has $\approx5$M parameters with a memory footprint of $59$ MB. We deploy the policy on a Jetson Orin integrated into our robot platform. At runtime, the policy executes at 0.5\,Hz and directly outputs the commanded light-intensity value.

Performance is evaluated on three autonomous runs. 
For each run, we collect two fixed-light baselines (0\% and 100\% intensity) as well as a run in which the policy controls the light. \Cref{tab:lightning_mast3rslam} summarizes the results. ILC consistently outperforms both baselines: in \texttt{113dark}, the 0\% and 100\% baselines fail at trajectory ratios of 0.26 and 0.44 (i.e., less than half the path), whereas ILC achieves 0.89. Similarly, in \texttt{kitchenloop}, the 0\% baseline fails at 0.48; although the 100\% baseline lasts longer, ILC outperforms both. Finally, in \texttt{start2corr}, all methods complete the path thanks to the less harsh environment, but ILC performs significantly better in localization error (WRMSE). \Cref{fig:placeholder} provides a qualitative comparison highlighting the limitations of fixed illumination. The 0\% baseline under-illuminates dark regions (a), while the 100\% baseline causes over-saturation on reflective surfaces (b). In contrast, ILC adapts its output to balance these competing effects across the sequence (c), maintaining task-consistent appearance.

%% file: tables/relight-static.tex
\begin{table}[t]
\centering
\small
\setlength{\tabcolsep}{4pt}
\renewcommand{\arraystretch}{1.15}
\caption{Relighting fidelity on paired-reference evaluations for custom static scenes. We report mean $\pm$ std over all evaluated frames, and additionally break down results by intensity transition.}
\label{tab:relight_fidelity_paired_meanstd_singlecol}

\begin{tabularx}{\columnwidth}{X c c c}
\toprule
Transition & \makecell{PSNR\\$\uparrow$} & \makecell{SSIM\\$\uparrow$} & \makecell{$\Delta$ Lum.\\(\%) $\downarrow$} \\
\midrule
All (mean $\pm$ std) & 27.34 $\pm$ 5.53 & 0.905 $\pm$ 0.020 & 11.5 $\pm$ 3.8 \\
25 $\rightarrow$ 50 & 25.81 $\pm$ 6.13 & 0.924 $\pm$ 0.018 & 11.6 $\pm$ 2.7 \\
50 $\rightarrow$ 100 & 28.53 $\pm$ 3.41 & 0.920 $\pm$ 0.014 & 2.5 $\pm$ 2.4 \\
100 $\rightarrow$ 50 & 25.49 $\pm$ 7.75 & 0.927 $\pm$ 0.028 & 9.9 $\pm$ 8.0 \\
75 $\rightarrow$ 25 & 27.56 $\pm$ 8.49 & 0.927 $\pm$ 0.034 & 8.9 $\pm$ 9.5 \\
50 $\rightarrow$ 0 & 29.13 $\pm$ 4.60 & 0.878 $\pm$ 0.026 & 21.5 $\pm$ 12.6 \\
100 $\rightarrow$ 0 & 27.52 $\pm$ 4.80 & 0.856 $\pm$ 0.034 & 14.5 $\pm$ 11.9 \\
\bottomrule
\hspace{5pt}
\end{tabularx}

\end{table}

%% file: tables/oracle-eval.tex
\begin{table*}[t]
\centering
\setlength{\tabcolsep}{2.5pt}
\renewcommand{\arraystretch}{1.05}
\caption{Oracle-derived Optimal Intensity Schedule (OIS) - SLAM performance}
\label{tab:ois_mast3rslam}

\begin{tabular*}{\textwidth}{@{\extracolsep{\fill}} l cc | cc | cc | cc}
\toprule
& \multicolumn{2}{c}{0\% baseline} 
& \multicolumn{2}{c}{OIS (oracle)} 
& \multicolumn{2}{c}{100\% baseline} 
& \multicolumn{2}{c}{OIS only} \\
\cmidrule(lr){2-3}\cmidrule(lr){4-5}\cmidrule(lr){6-7}\cmidrule(lr){8-9}
Sequence
& \makecell{Trajectory Ratio\\$C\uparrow$} & \makecell{$\mathrm{WRMSE}$\\$\downarrow$}
& \makecell{Trajectory Ratio\\$C\uparrow$} & \makecell{$\mathrm{WRMSE}$\\$\downarrow$}
& \makecell{Trajectory Ratio\\$C\uparrow$} & \makecell{$\mathrm{WRMSE}$\\$\downarrow$}
& \makecell{Light\\$\mu$ (\%)} & \makecell{Power\\$P$ (W)} \\
\midrule
\texttt{101backdoor} & \textbf{0.97} & 0.684 & 0.86 & \textbf{0.310} & 0.87 & 0.323 & 62 & 22.9 \\
\texttt{113through} & \red{0.04} & 18.493 & \textbf{0.50} & \textbf{0.944} & \red{0.04} & 13.736 & 88 & 28.4 \\
\texttt{203} & \red{0.37} & 1.728 & \textbf{0.46} & \textbf{1.000} & \red{0.42} & 7.495 & 19 & 13.5 \\
\texttt{230a} & 0.60 & 1.133 & \textbf{0.71} & \textbf{0.704} & 0.68 & 0.731 & 14 & 12.5 \\
\texttt{corridor2lab} & \textbf{1.00} & 0.593 & 0.99 & \textbf{0.372} & 0.99 & 0.462 & 85 & 27.8 \\
\texttt{entry2corr} & 0.69 & 2.627 & 0.69 & \textbf{1.153} & \textbf{0.70} & 1.558 & 91 & 29.0 \\
\texttt{lab2service} & 0.94 & 0.507 & 0.98 & \textbf{0.292} & \textbf{0.99} & 0.300 & 90 & 28.8 \\
\texttt{labin} & \red{0.51} & 0.631 & \textbf{0.98} & \textbf{0.447} & \red{0.53} & 0.850 & 74 & 25.3 \\
\texttt{tunnelin} & \red{0.38} & 1.951 & \textbf{0.81} & \textbf{0.604} & 0.80 & 0.635 & 94 & 29.6 \\
\texttt{tunnelout} & 0.97 & 0.857 & \textbf{0.99} & \textbf{0.642} & 0.97 & 0.817 & 70 & 24.5 \\
\bottomrule
\end{tabular*}

\vspace{0.5mm}
\footnotesize{
\textbf{Notes:} Trajectory Ratio $C=1 - \frac{|T_{\text{pred}} - T_{\text{gt}}|}{T_{\text{gt}}}$.
Weighted RMSE: $\frac{\mathrm{ATE\ RMSE}}{C^{2}}$.
Power derived from $P = 21.3636 \cdot I + 9.5455$.}
\end{table*}

%% file: tables/rollout.tex






\begin{table*}[t]
\centering
\scriptsize
\setlength{\tabcolsep}{3pt}
\renewcommand{\arraystretch}{1.05}
\caption{Active Illumination Control using \qq{} – SLAM Performance (RMSE removed).}
\label{tab:lightning_mast3rslam}

\begin{tabular*}{\textwidth}{@{\extracolsep{\fill}} l cc | cc | cc | cc}
\toprule
& \multicolumn{2}{c}{0\% baseline}
& \multicolumn{2}{c}{\qq{} (Online)}
& \multicolumn{2}{c}{100\% baseline}
& \multicolumn{2}{c}{\qq{} only} \\
\cmidrule(lr){2-3}\cmidrule(lr){4-5}\cmidrule(lr){6-7}\cmidrule(lr){8-9}

Sequence
& \makecell{Trajectory Ratio\\$C\uparrow$}
& \makecell{$\mathrm{WRMSE}$\\$\downarrow$}
& \makecell{Trajectory Ratio\\$C\uparrow$}
& \makecell{$\mathrm{WRMSE}$\\$\downarrow$}
& \makecell{Trajectory Ratio\\$C\uparrow$}
& \makecell{$\mathrm{WRMSE}$\\$\downarrow$}
& \makecell{Light\\$\mu$ (\%)}
& \makecell{Power\\$P$ (W)} \\
\midrule
\texttt{113dark} 
& \red{0.26} & 2.756
& \textbf{0.89} & 1.405
& \red{0.44} & \textbf{0.641}
& 48.28 & 19.86 \\

\texttt{kitchenloop} 
& \red{0.48} & 0.744
& \textbf{0.91} & \textbf{0.393}
& 0.88 & 0.425
& 60.42 & 22.45 \\

\texttt{start2corr} 
& 0.95 & 0.220
& \textbf{0.99} & \textbf{0.123}
& 0.98 & 0.249
& 47.33 & 19.66 \\
\bottomrule
\end{tabular*}

\vspace{0.5mm}
\end{table*}

%% file: sections/disc.tex
\section{Discussion}
\label{sec:discussion}



\subsection{Effect of Exposure Control}
Camera auto-exposure (AE) algorithms are typically designed to produce \emph{neutral-brightness} images i.e., images that are neither under- nor over-exposed according to photometric heuristics. This goal is, in general, not aligned with ours: we aim to maximize downstream SLAM performance. In challenging conditions, a neutrally exposed image can be suboptimal for tracking,for example, during bright-dark transitions, intentionally sacrificing highlights may preserve mid-tone gradients and improve feature consistency across frames.

To isolate the effect of active illumination from these potentially conflicting objectives, we fix the camera exposure throughout all experiments. We choose this fixed setting to be approximately neutral \emph{on average} across our data, ensuring that neither baseline illumination (0\% and 100\% light) is systematically disadvantaged by exposure choice. Joint control of active illumination and exposure is also feasible within our framework. In particular, the oracle and learned policy could be extended to operate over a two-dimensional action space (illumination, exposure), yielding a 2D search/inference problem. We leave this extension to future work.




\subsection{Oracle Objective Design}

The oracle’s objective is intentionally constructed as a proxy for downstream SLAM performance, since using SLAM itself as the optimization objective is non-trivial. In particular, RMSE error from SLAM is not a local function of $I_t, I_{t+1}$ or intensity $k_t$, the final trajectory quality depends on long-horizon phenomena such as multi-frame data association, loop closures, and discrete failure modes (e.g., tracking loss and subsequent relocalization). Moreover, when tracking fails, many SLAM pipelines return undefined or discontinuous signals (and effectively no usable gradient), which makes the objective ill-suited for per-step supervision and brittle as a term inside dynamic programming (DP), where costs must be well-defined for every candidate state and transition.

To enable stable offline optimization, we instead define 
a proxy energy that captures the dominant factors influencing SLAM robustness, while treating end-to-end SLAM performance as a second-order objective evaluated outside the dynamic programming loop.


\subsection{Limitations}
The quality of the oracle supervision is bounded by the fidelity of CLID relighting: OIS is optimized over \emph{synthesized} candidate observations, and any relighting error can bias the resulting schedule and consequently, the imitation policy. While CLID relights our indoor operating conditions effectively, we expect reduced generalization under distribution shift (e.g., outdoor scenes with sunlight and higher dynamic range or novel materials). Further, if the input images are not RAW (i.e., they have undergone ISP processing such as gamma, tone mapping, and clipping), the resulting non-linearities break the scene-linear image formation model required by the CLID-network, and the synthesized candidates will have poor image quality. \kar{However, we have extensively tested CLID in indoor scenarios (\Cref{tab:relight_fidelity_paired_meanstd_singlecol}) and have found it to be fairly representative of low-lit indoor scenes in a building. }

\vspace{-0.2cm}



%% file: sections/conc.tex
\section{Conclusion}
\label{sec:conclusion}

In this paper we introduce a novel active illumination control framework. The three-stage pipeline includes an image relighting network (CLID) for data generation and expansion, a sequence-level oracle expert which solves for an optimal intensity schedule, balancing image utility, feature matching, power and smoothness, and an imitation learning-based controller that is supervised by the oracle. The resulting controller, ILC, runs onboard a robot in real-time and commands light-intensity values to an onboard, co-located light. Across various trajectories, our approach yields more reliable SLAM performance compared to fixed illumination baselines while reducing unnecessary power usage, demonstrating that task-aware active illumination can be solved and deployed and is a practical way to improve perception quality. Future work includes coupling illumination with active exposure and extending the approach to broader environments.